\documentclass[runningheads]{llncs}
\usepackage[T1]{fontenc}
\usepackage{subcaption}
\usepackage{algorithm}

\usepackage{graphicx}
\usepackage[para]{footmisc}

\begin{document}
%

\title{Beyond the Hype: Benchmarking LLM-Evolved Heuristics for Bin Packing}
%

%
\author{Kevin Sim\orcidID{0000-0001-6555-7721} \and Quentin Renau\orcidID{0000-0002-2487-981X} \and 
Emma Hart\orcidID{0000-0002-5405-4413}}
\authorrunning{Sim et. al.}
%
\institute{Edinburgh Napier University\\
\email{\{k.sim, q.renau,e.hart\}@napier.ac.uk}}
\maketitle              
\begin{abstract}
Coupling Large Language Models (LLMs) with Evolutionary Algorithms has recently shown significant promise as a technique to design new heuristics that outperform existing methods, particularly in the field of combinatorial optimisation. An escalating arms race is both rapidly producing new heuristics and improving the efficiency of the processes evolving them. However, driven by the desire to quickly demonstrate the superiority of new approaches, evaluation of the new heuristics produced for a specific domain is often cursory: testing on very few datasets in which instances all belong to a specific class from the domain, and on few instances per class.  Taking bin-packing as an example, to the best of our knowledge we conduct the first rigorous benchmarking study of new LLM-generated heuristics, comparing them to well-known existing heuristics across a large suite of benchmark instances using three performance metrics.  For each heuristic, we then evolve new instances `won’ by the heuristic and perform an instance space analysis to understand where in the feature space each heuristic performs well. We show that most of the LLM heuristics do not generalise well when evaluated across a broad range of benchmarks in contrast to existing simple heuristics, and suggest that any gains from generating very specialist heuristics that only work in small areas of the instance space need to be weighed carefully against the considerable cost of generating these  heuristics.

\end{abstract}
\section{Introduction}
Designing heuristics for solving optimisation problems  can be a time-consuming process, often requiring expert knowledge. Automating the design of heuristics (ADH) is thus an appealing direction for research. Multiple approaches to this have been proposed over the last two decades. For example, the field of hyper-heuristics encompasses methods that either select the best performing heuristic from a pre-defined set or generates new heuristics by recombining components of existing heuristics~\cite{burke2013hyper,sanchez2020systematic} that have shown particular promise on combinatorial optimisation:  approaches based on genetic programming in particular have found success here~\cite{burke2012automating,djurasevic2022automated}. Other approaches use the configuration tool \textit{irace}~\cite{irace} to select individual components of a heuristic or algorithm~\cite{martin2024automatic}, while methods that use machine-learning techniques are also gaining in popularity~\cite{bengio2021machine}.

With the advent of Large Language Models (LLMs), new attempts to automate the design of heuristics have been proposed, using an LLM as a generator of new heuristics. Unlike the previous methods described which are limited by a heuristic space predefined by human experts, LLMs have access to vast amounts of information, as well as being able to search directly and efficiently in the space of code  and/or linguistic representations of a heuristic. The release of FunSearch (`searching in the function space')~\cite{romera2024mathematical} demonstrated that an evolutionary procedure that paired a pre-trained LLM with a systematic evaluator was able to discover new heuristics that improved on long established baselines in two combinatorial domains, suggesting that ``\textit{it is possible to make discoveries for established open problems using LLMs}'', with improved frameworks quickly following~\cite{liu2024evolution,ye2024reevo}.

Despite the promising results reported, to date, it appears that these results have not been subject to deep interrogation. Taking online bin-packing\footnote{items arrive one at a time and must be packed immediately; no information is available as to what items will arrive in future} as an exemplar domain given that it has recently received much attention~\cite{romera2024mathematical,liu2024evolution,van2024loop}, several gaps in analysis are apparent.  Firstly, the success of an LLM technique is judged by evolving heuristics for very few representative classes from the domain.   
It is clear that a heuristic evolved using one of the LLM frameworks will specialise to the class of instances used during evolution. However, the extent to which the evolved heuristics might generalise across other classes has never been investigated. Given that the hand-designed heuristics used for performance comparison are well-known to generalise well across many classes~\cite{johnson1973near}, it is interesting therefore to examine the trade-off in the specialist vs generalist nature of these heuristics by evaluating their performance across many datasets. This is particularly important given the considerable computational cost of generating a heuristic with an LLM\footnote{It is estimated the price of a FunSearch experiment running on Google Cloud is between  $800\$ - 1400\$$, with energy usage in the region of $250$ – $500$ kWh~\cite{romera2024mathematical}}.

Furthermore, in the optimisation literature in general, it is untypical to benchmark a new algorithm on very
few classes or indeed few instances per class. However, the heuristics from the EoH~\cite{liu2024evolution} framework are only evaluated on instances from a single Weibull distribution~\cite{rinne2008weibull}, with five instances evaluated from six variants of this class in which the bin capacity and number of items  are varied (resulting in $30$ instance in total). One of the FunSearch heuristics is evaluated using four datasets from the literature~\cite{beasley1990or}, but each dataset has instances in which items are drawn from the \textit{same} uniform distribution, and there are only $20$ instances per class.

In contrast, a recent evaluation of TSP solvers~\cite{mcmenemy2019rigorous} benchmarked against $1{,}800$ TSP instances from six different families were evaluated.
$9{,}720$ instances are available in a benchmark suite for the travelling thief problem~\cite{polyakovskiy2014comprehensive}, while in continuous optimisation, the BBOB benchmark~\cite{cocoJournal} provides $24$ functions with a range of properties and from which multiple instances can be generated per function, and is widely used by the community. Finally, while it is reasonable to limit comparison to the best known and most general bin-packing heuristics Best-Fit and First-Fit~\cite{johnson1973near}, there are other established online heuristics that perform well on certain types of instance, hence a rigorous benchmarking exercise should also include these heuristics.

We therefore conduct an in-depth  benchmarking exercise in the online bin-packing domain of the most recent LLM-generated heuristics.  Our contributions are as follows:

\begin{itemize}
\item We evaluate five LLM-generated heuristics on \textit{$6{,}064$} instances obtained from \textit{$12$}  datasets drawn from the wider literature, each with with different characteristics, in order to understand trade-offs in generality vs speciality. 

\item We compare the LLM-generated heuristics to five hand-designed heuristics on each of the $12$ datasets, extending the number of hand-designed heuristic solvers used for comparison to five, and measuring performance against three different evaluation metrics. 

\item We tune the hard-coded constants in the LLM-evolved heuristics to understand how robust they are to the evolved parameters and to determine whether they can be improved by a standard tuning technique.

\item To gain insights into what part of the instance space is  covered by each of LLM-evolved heuristics, we evolve $100$ new instances won by five LLM-heuristics and five hand-designed heuristics and use Instance Space Analysis~\cite{smith2023instance} to illustrate coverage of the space in 2d.
\end{itemize}

\section{Background}

As already noted, there has  been a recent rapid proliferation of research leveraging LLMs as algorithm generators. Many diverse examples now exist. Pluhacek {\em et. al.}~\cite{pluhacek2024using} use an LLM to iteratively enhance a selected baseline metaheuristic, showing that  significant improvements can made, albeit not consistently. 
Van Stein{\em et. al.}~\cite{van2024llamea}  show that an LLM framework $LLaMEA$ can be used to generate novel black-box metaheuristic optimisation algorithms automatically,   generating multiple algorithms that outperform state-of-the-art optimisation algorithms  such as CMA-ES~\cite{HansenO01} and Differential Evolution~\cite{StornP97} on the five-dimensional black-box optimisation benchmark (BBOB)~\cite{cocoJournal}. 

Focusing on the generation of heuristics for well-known combinatorial optimisation problems, FunSearch~\cite{romera2024mathematical}  is an evolutionary method that iteratively feeds selected heuristics described purely in code to an LLM which is used to generate new samples. Results in the 1d online bin-packing domain show that a heuristic evolved  by FunSearch on $5$ training instances drawn from the OR1 benchmark dataset~\cite{beasley1990or} outperforms the best known hand-designed heuristics on unseen instances from OR1 as well as instances from each three other OR datasets~\cite{beasley1990or}, in which items are drawn from the same distribution but the number of items varies. The FunSearch framework was also used to evolve a heuristic applicable to  bin-packing instances where item sizes are drawn from a Weibull distribution (generally recognised as more closely following the distributions found in many real-world scheduling problems~\cite{castineiras2012weibull}). Again, the evolved heuristic was demonstrated to outperform the best known hand-designed heuristics  according to the metric 'percentage of bins over optimal' on three different Weibull datasets each containing $5$ instances, in which each dataset had a different number of items.

Extending the FunSearch paradigm, a framework Evolution of Heuristics (EoH) was recently proposed~\cite{liu2024evolution}. EoH simultaneously evolves `thoughts' (representing the key concepts of a heuristic in natural language)  \textit{and} code (an executable implementation of a heuristic). This contrasts to FunSearch which only searches in the space of code. The approach is shown to be both
effective and efficient in generating high-performance heuristics: specifically in  the domain of online bin-packing it generates heuristics that outperform hand-designed heuristics \textit{and} FunSearch on various test sets of instances with items drawn from a Weibull distribution with varying number of items/bin capacity, at a fraction of the computational cost. As in~\cite{romera2024mathematical} however, test sets all had instances with item sizes drawn from the same distribution, and comprise of only five instances, so despite the promising results reported, it is unclear how the evolved heuristics might generalise or perform outside of the settings they were evolved for.

Despite this stream of promising results,  some questions are beginning to be raised regarding the benchmarking of evolutionary LLM approaches. Zhang {\em et. al.}~\cite{zhang2024understanding} point to weaknesses in evaluation methodologies with respect to the choice of LLMs and their various parameter settings; the comparison of evolved algorithms or heuristics to inadequate baselines (e.g. random search), and  finally the lack of understanding of the contribution of individual components of frameworks to overall performance. They propose a new baseline LLM method to which results from new frameworks can be compared, but do not specifically tackle the issues of generality vs speciality, the diversity in datasets used or the number of instances evaluated. They conclude with the statement ``more diverse benchmarks and applications are needed to establish a better understanding of this emergent paradigm for AHD''. 

We directly address this comment by  rigorously benchmarking  $5$ LLM-generated heuristics  (1) on datasets with a large number of instances and (2) across diverse datasets in the bin-packing domain, and then  use instance space analysis (ISA)~\cite{smith2023instance} to shed light in the strengths and weaknesses of each heuristic across an instance space.

\section{Benchmarking}
We first conduct a benchmarking study in which we apply LLM-evolved heuristics and hand-designed heuristics from the literature to a large set of benchmarks. The heuristics and datasets used are described below.

\subsection{Heuristics}
\label{sec:LLMheuristics}
Five common hand-designed heuristics first described in~\cite{johnson1973near} are evaluated:

\begin{itemize}
\item \textbf{Next-Fit} (NF) always keeps a single open bin. When the new item does not fit into it, it closes the current bin and opens a new bin.

\item \textbf{First-Fit} (FF) keeps all bins open, in the order in which they were opened. It attempts to place each new item into the first bin in which it fits. 

\item \textbf{Best-Fit }(BF) keeps all bins open. It places each new item into the bin with the maximum load in which it fits

\item \textbf{Worst-Fit} (WF) attempts to place each new item into the bin with the minimum load. 

\item \textbf{Almost Worst-Fit} (AWF) attempts to place each new item inside the second most empty open bin (or emptiest bin if there are two such bins). If it does not fit, it tries the most empty one.
\end{itemize}

\begin{figure}
    \centering
    \includegraphics[width=\linewidth]{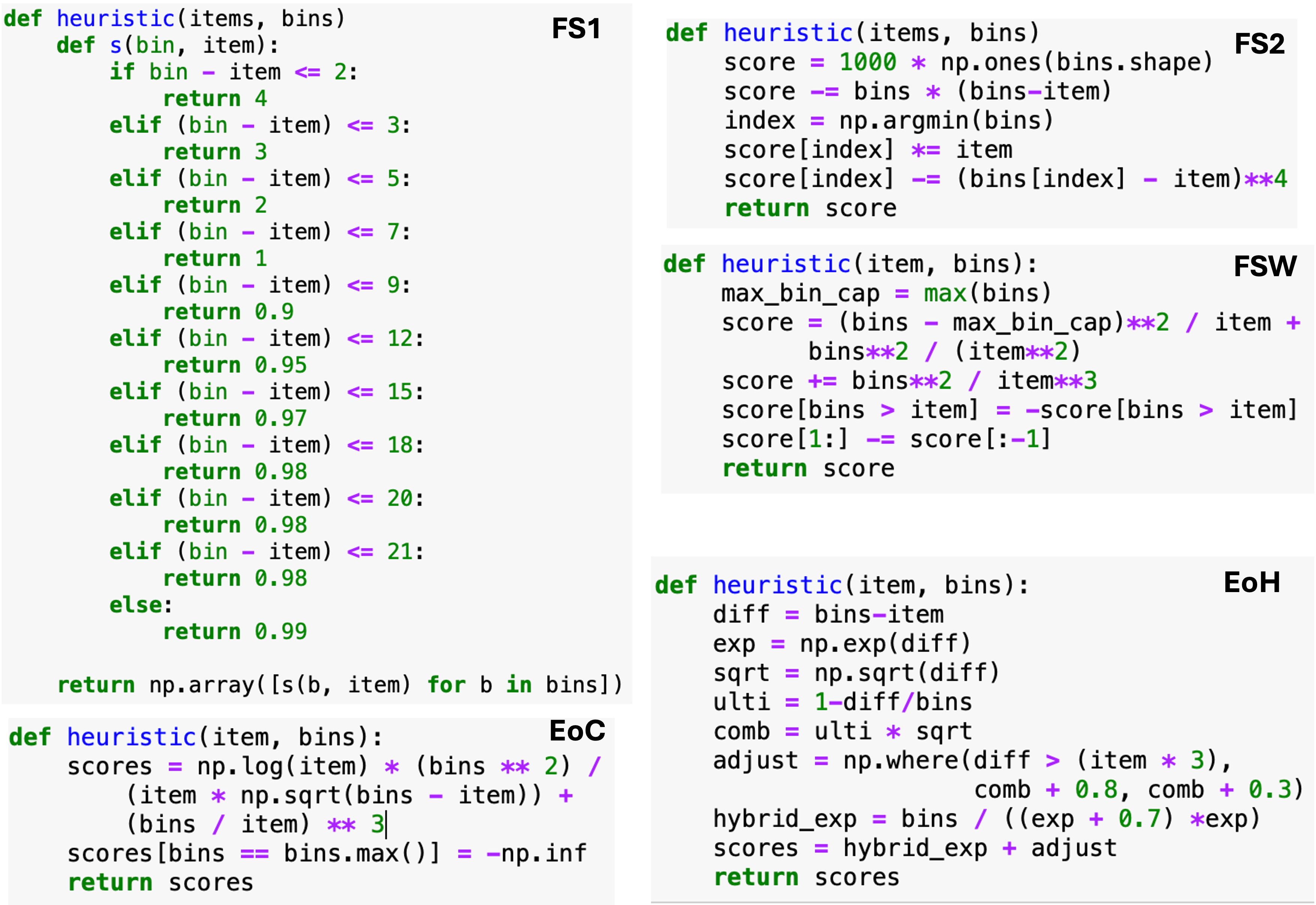}
    \caption{Code of LLM-evolved heuristics trained. FS1, FS2, FSW are adapted from code provided in~\cite{romera2024mathematical}, $EoC$ and $EoH$ from code provided in~\cite{liu2024evolution}.}
    \label{fig:allheuristics}
\end{figure}

We also evaluate $5$ LLM-generated heuristics, defined in code in Figure~\ref{fig:allheuristics}.
We use the skeleton in the Juypter notebook provided by~\cite{romera2024mathematical} to run the FunSearch evolved heuristics. The heuristics evolved in~\cite{liu2024evolution} have the same form as those evolved by FunSearch and can be run in the same notebook. Note that we have named the FunSearch heuristics ourselves in order to more easily refer to them through the text.

\begin{itemize}
    \item $\textbf{FS1}$ : Evolved using FunSearch~\cite{romera2024mathematical}  using $20$ training instances
    generated at random from the same parameters as instances in the first of the OR-Library bin packing benchmarks, OR1~\cite{beasley1990or}. The heuristic is essentially a list of \textit{if/then} statements of the form `\textit{if (remainingCapacity-itemsize} $\le x$ \textit{then score} $=y$'. The evolutionary process defines $10$ integer values for $x$, and 10 floating-point values for $y$.

    \item $\textbf{FS2}$: Also evolved using FunSearch~\cite{romera2024mathematical} using the same training instances as described for $FS1$. The heuristic is a function that penalises bins with large capacities and bins where the fit is not tight. There are two integer parameters in this function (chosen by the evolutionary process),  specifying an initial penalty value and  scaling factor.

    \item $\textbf{FSW}$: Evolved using FunSearch~\cite{romera2024mathematical} using training instances from a Weibull distribution~\cite{castineiras2012weibull}. The score for each bin is the sum of three terms creating a scoring function dependent on item size and bin capacity. There are $5$ integer values (chosen by the evolutionary process) in this function used to raise the item size or bin capacity to the power $i$.
    
    \item $\textbf{EoH}$: Evolved using EoH~\cite{liu2024evolution} using a Weibull dataset for training as with $FSW$.
    The evolutionary framework evolves
    both linguistic descriptions of a heuristic \textit{and} executable code. The best heuristic evolved  incorporates a weighted average of the utilisation ratio, dynamic adjustment, and an exponentially decaying factor, with different parameter settings to minimise the number of used bins. It contains $4$ floating-point parameter values (chosen by the evolutionary process).

    \item $\textbf{EoC}$: 
    Also from~\cite{liu2024evolution}, here the evolutionary framework searches for heuristics entirely in the \textit{code} space. The best function    
    evolved is similar to the evolved $FS2$ heuristic in its structure. $EoC$ contains two integer parameters chosen by the evolutionary process.
\end{itemize}

Results presented in~\cite{liu2024evolution} show that $EoH$ is superior to  $EoC$, and outperforms $FSW$ on four out of $6$ datasets, equals $FSW$ on one dataset and loses on one. $EoH$ is therefore considered a stronger heuristic than $FSW$.
For the $FS1$ and $FS2$ heuristics evolved using FunSearch, results in~\cite{romera2024mathematical} show that $FS1$ outperforms hand-designed heuristics on the four datasets in the OR-Library~\cite{beasley1990or}. Results for $FS2$ are not shown in~\cite{romera2024mathematical}; we therefore ran the heuristic code ourselves on the same four OR datasets~\cite{beasley1990or}, finding that $FS1$ outperforms $FS2$ on ${OR2, OR3, OR4}$ and loses only by a small margin on $OR1$.

\subsection{Benchmark Datasets}
\label{sec:dataset}
A diverse range of benchmark datasets drawn from the literature are used for comparisons. The majority of datasets are downloaded from the bin-packing library BPPLib~\cite{bpplib}

Two datasets are taken from the ORLibrary~\cite{beasley1990or}. We exclude the datasets `binpack1' and 'binpack2' also available for the ORlibrary from our benchmarking exercise as instances from distributions identical to those used to generate these datasets are used in the training and validation procedures of the LLM frameworks used to evolve heuristics. The final dataset in the table was proposed by~\cite{angelopoulos2023online} and downloadable from~\cite{angelopGithub}: it uses item sizes drawn from a Weibull distribution. 

Some of these datasets were originally created to evaluate \textit{offline} packing algorithms, therefore the order that items are described per instance in the dataset represents the optimal ordering. To use the instances in an \textit{online} scenario, we follow the method used by Angelopoulos {\em et. al.}~\cite{angelopoulos2023online} and randomly shuffle the items in each dataset. The new shuffled datasets are provided in a repository accompanying the paper~\cite{dataLLMBP}.

\begin{table}
    \centering
    \caption{Benchmark datasets:  items in instances in the datasets indicated in \textit{italics} are packed in the order given in the dataset. For the remaining datasets  which were originally designed for offline packing,  item sizes are randomly shuffled. The shuffled instances are provided at~\cite{dataLLMBP}.}
    \label{tab:datasets}
    \begin{tabular}{|c|c|c|c|c|c|}
    \hline
       Name  &  Num Instances & Capacity (C)  & Items  &  Source\\
       \hline
       AugmentedIRUP ~\cite{delorme2016bin}& $250$ &  \{2.5K-80K\} & \{201-1002\}  & BPPLib \\
    AugmentedNonIRUP~\cite{delorme2016bin}& $250$ & \{2.5K-80K\}  &  \{201-1002\}  & BPPLib \\
        Irnich~\cite{gschwind2016dual} & $240$   & \{500K, 1500K \} & [(2/15)*C-(2/3)*C] & BPPLib \\
     
       Schoenfield\_Hard  & $28$ & $10000$  &  $\{160$-$200\}$ & BPPLib \\
       Scholl\_1~\cite{scholl1997bison}  & $720$  & $\{100$-$150\}$  & $500$ &BPPLib  \\
       Scholl\_2~\cite{scholl1997bison} & $480$ & $1000$ & $500$  & BPPLib \\
       Scholl\_Hard ~\cite{scholl1997bison} & $10$ & $100K$ & $200$ &BPPLib  \\
       Schwerin\_1~\cite{scholl1997bison}&  $100$ & $1K$  &$100$  &BPPLib  \\
      Schwerin\_2~\cite{scholl1997bison}& $100$  & $1K$ &  $120$ & BPPLib  \\
       Waescher~\cite{wascher1996heuristics}& $17$  & $10K$ & \{27-239\}  &  BPPLib\\
           \textit{Randomly Generated}   & $3{,}840$ & $\{50$-$1K\}$ &   $\{50-1K\}$ & BPPLib\\
          \textit{Angelopoulos}~\cite{angelopoulos2023online}& $29$ & $100$ &  $10^6$ &~\cite{angelopGithub}\\
           
          \hline
    \end{tabular}
\end{table}

\section{Benchmarking Results}
\label{sec:BenchmarkResults}

We apply each of the $10$ heuristics to each of the $12$ benchmark datasets described in Section~\ref{sec:dataset}. All heuristics are deterministic so are only run once. We score each heuristic on each dataset according to three metrics: 

\begin{enumerate}
\item \textbf{Average Excess Bins (AEB)}: The mean percentage of bins over optimal across the dataset, where optimal is calculated according to a lower bound defined as the sum of the $n$ item sizes divided by bin capacity $C$, i.e.  $\sum_{i=1}^{i=n}  size(i)/C$.
\item \textbf{Falkanauer fitness}: defined in~\cite{falkenauer1992genetic} this prefers `well-filled' bins rather than a set of equally filled bins. This is calculated as the average, over all bins, of the $k^{th}$ power of 'bin efficiency', i.e. $\sum_{i=1}^{i=n}\frac{(fill_i/C)^k}{n}$.
\item \textbf{Wins}: the percentage of instances in a dataset won by a heuristic. A heuristic wins an instance if it produces a solution with the minimum number of bins compared to the remainder of the portfolio\footnote{a `win' is counted if the heuristic result is better or equal to the best value}. 
\end{enumerate}

\begin{figure}[h]
    \centering
    \begin{subfigure}[t]{0.5\textwidth}
        \centering
        \includegraphics[width=\textwidth]{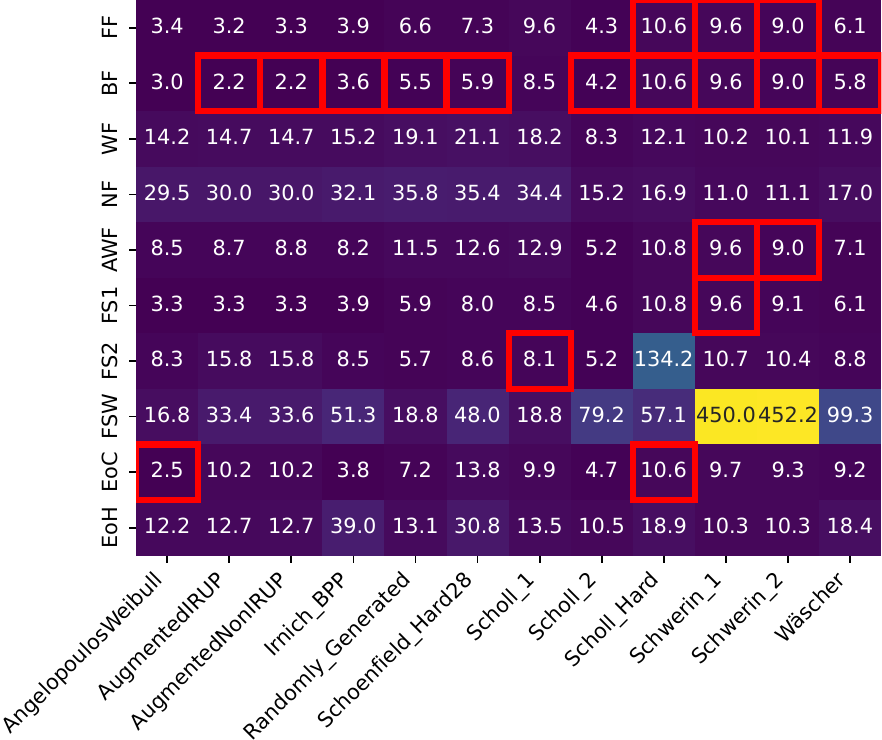}
        \caption{Average Excess Bins}
        \label{fig:heatmaps_bins}
    \end{subfigure}%
    ~ 
    \begin{subfigure}[t]{0.5\textwidth}
        \centering
        \includegraphics[width=\textwidth]{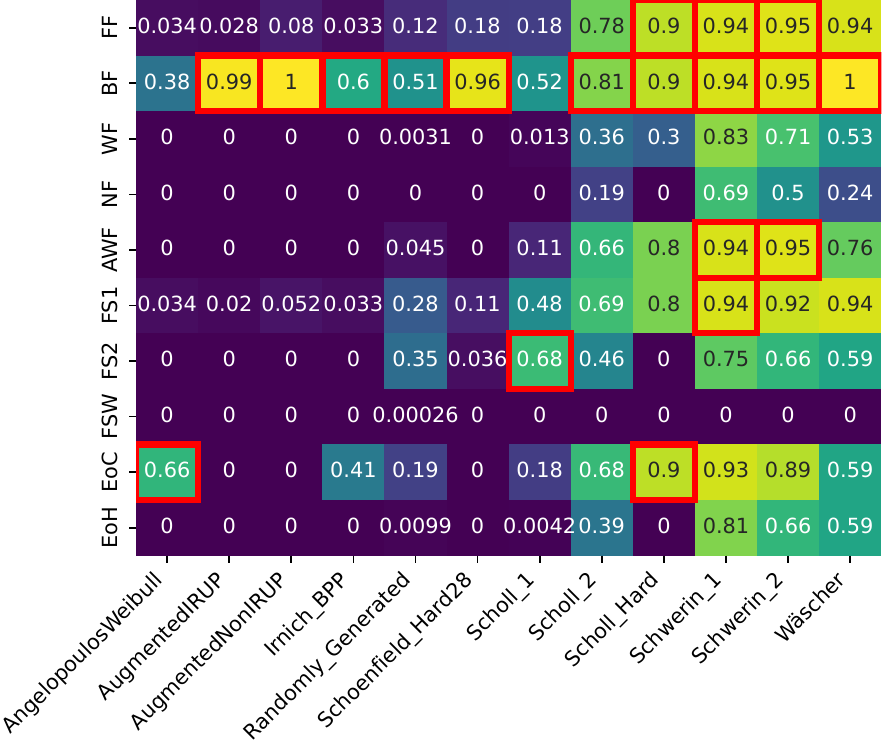}
        \caption{Wins on bins}
        \label{fig:heatmaps_wins}
    \end{subfigure}
    ~
     \begin{subfigure}[t]{0.45\textwidth}
        \centering
        \includegraphics[width=\textwidth]{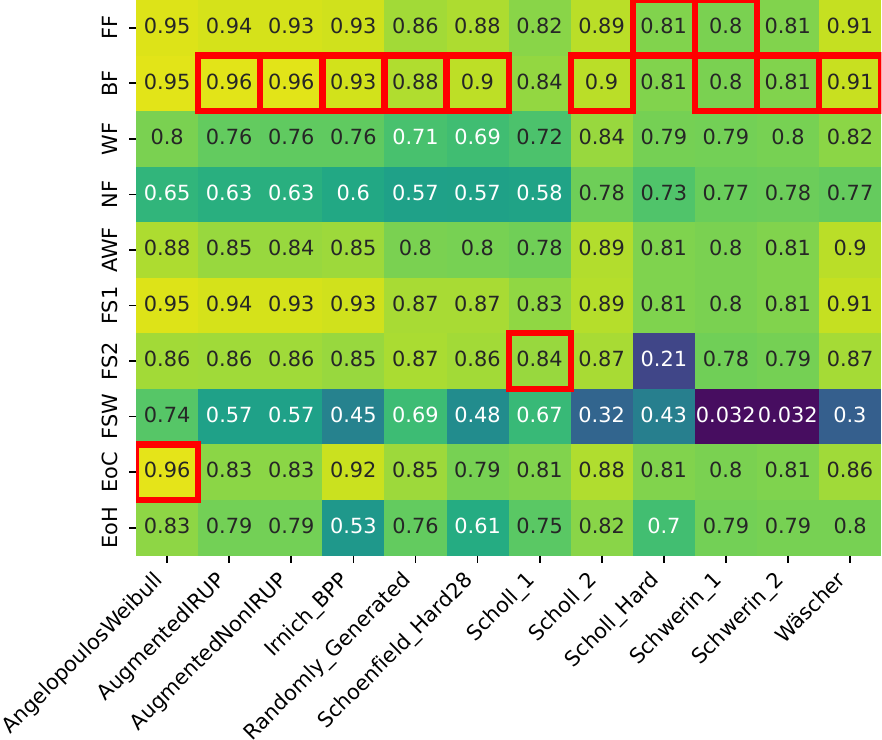}
        \caption{Falkanauer}
        \label{fig:heatmaps_F}
    \end{subfigure}
    ~
    \begin{subfigure}[t]{0.5\textwidth}
        \centering
        \includegraphics[width=\textwidth]{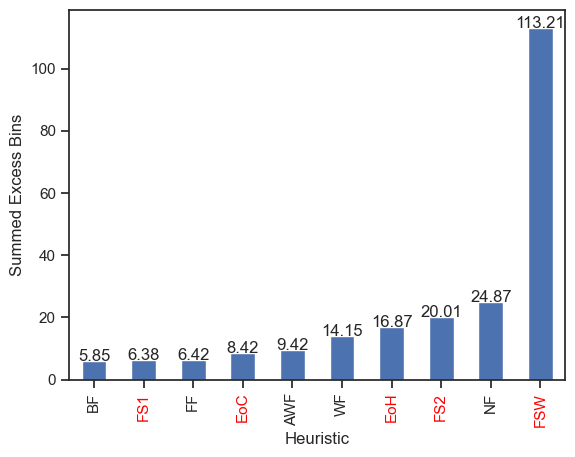}
        \caption{Average Excess Bins over all datasets}
        \label{fig:summary}
    \end{subfigure}
    \caption{\label{fig:heatmaps} (a-c): Heatmaps showing heuristic performance per dataset for 3 metrics: average excess bins; Falkanauer fitness; percentage of instances won. Red boxes highlight the best values by column. (d): Histogram showing the summed excess bins per heuristic over all datasets.}
\end{figure}

Results are first presented in the form of three heatmaps (Figure~\ref{fig:heatmaps}(a-c)) showing the value of each  metric per dataset and heuristic.  Figure~\ref{fig:heatmaps}(d) shows the sum of the AEB metric over all datasets in the benchmark, ranked in order of superiority. The hand-designed heuristic $BF$ ranks 1st, shown in  Figure~\ref{fig:summary}: in fact this heuristic wins on all but $2$ datasets, Weibull and Scholl-1 (Figure~\ref{fig:heatmaps_bins}). The LLM heuristic $FS1$ ranks second overall, although it does not win outright on any dataset. It ties for 1st place with three hand-designed heuristics on the Schwerin-1 dataset, and is close to $BF$ on most other datasets.  On the other hand, the LLM heuristic $FS2$ \textit{is} the outright winner on one dataset (Scholl-1) but ranks in 8th place overall, showing that it has a role as a specialist heuristic but is not a good generalist.

$EoC$ (from~\cite{liu2024evolution}) ranks 4th overall and is the best heuristic for the Angelopolous Weilbull dataset. It ties for first place on Scholl-Hard (with $BF$ and $FF$). This heuristic was evolved using data from a Weibull distribution during training (although with different parameters to the Angelopolous Weibull dataset), hence it is not surprising that it performs well here. Note that it performs better than $EoH$ on this suite, in contrast to results in~\cite{liu2024evolution}  which show that $EoC$ is always outperformed by $EoH$ on Weibull datasets.  $EoH$ never wins according to the $AEB$ metric on any dataset, but is never the worst performer. Figure~\ref{fig:heatmaps_wins} shows however that it provides the best or equal best solution on $81\%$ of the instances in the Schwerin-1 dataset but when results are averaged across the whole data it loses out to $\{BF,FF,AWF,FS1\}$, with $AEB=10.3$ vs $AEB=9.6$ for the winning heuristics.

The FunSearch heuristic $FSW$ --- evolved using instances from a Weibull distribution ---  is the worst  performing  heuristic on every dataset with respect to the AEB metric, often by a large margin and in fact only wins a single instance from a single dataset (`Randomly-Generated') (Figure~\ref{fig:heatmaps_wins}. Two of the hand-designed heuristics ($NF$ and $WF$) do not win (or draw) on any of the datasets but do win some instances per dataset.

The patterns just described are mirrored in the heatmap plotted using the Falkanauer metric (Figure~\ref{fig:heatmaps_F}) suggesting that  heuristics that provide good packing in terms of AEB also optimise the Falkanauer metric which minimises empty space. We omit plots on this metric for this reason in the remainder of the paper.

\begin{figure}[ht!]
    \centering
    \begin{subfigure}[t]{0.50\textwidth}
        \centering
        \includegraphics[height=1.9in]{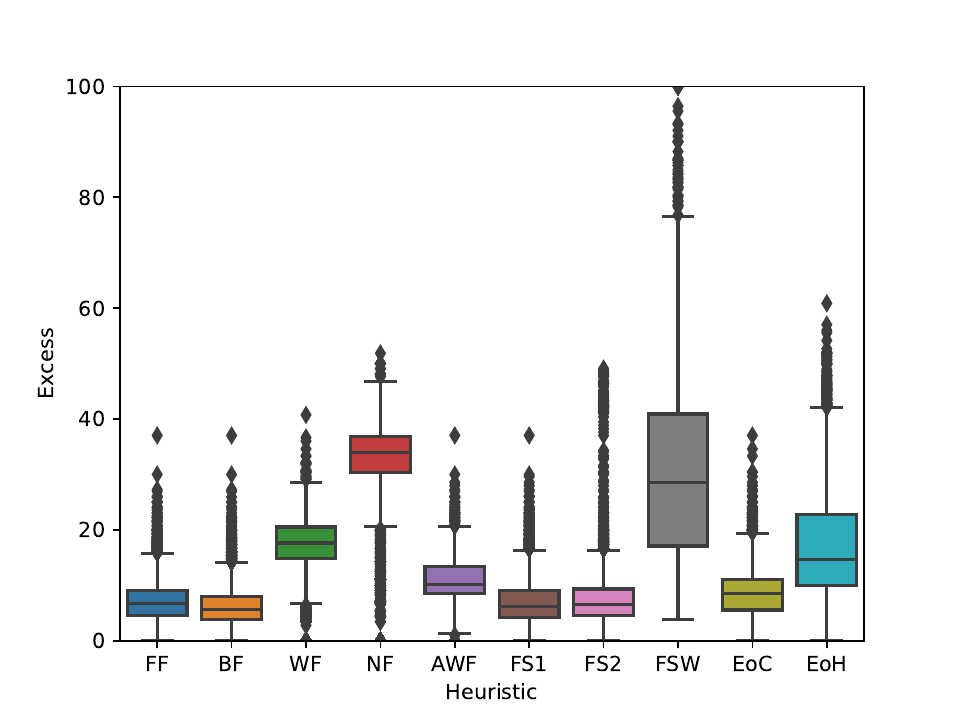}
        \caption{Average Excess Bins, capped at $100\%$}
        \label{fig:boxplots_bins}
    \end{subfigure}%
    ~ 
    \begin{subfigure}[t]{0.50\textwidth}
        \centering
        \includegraphics[height=1.9in]{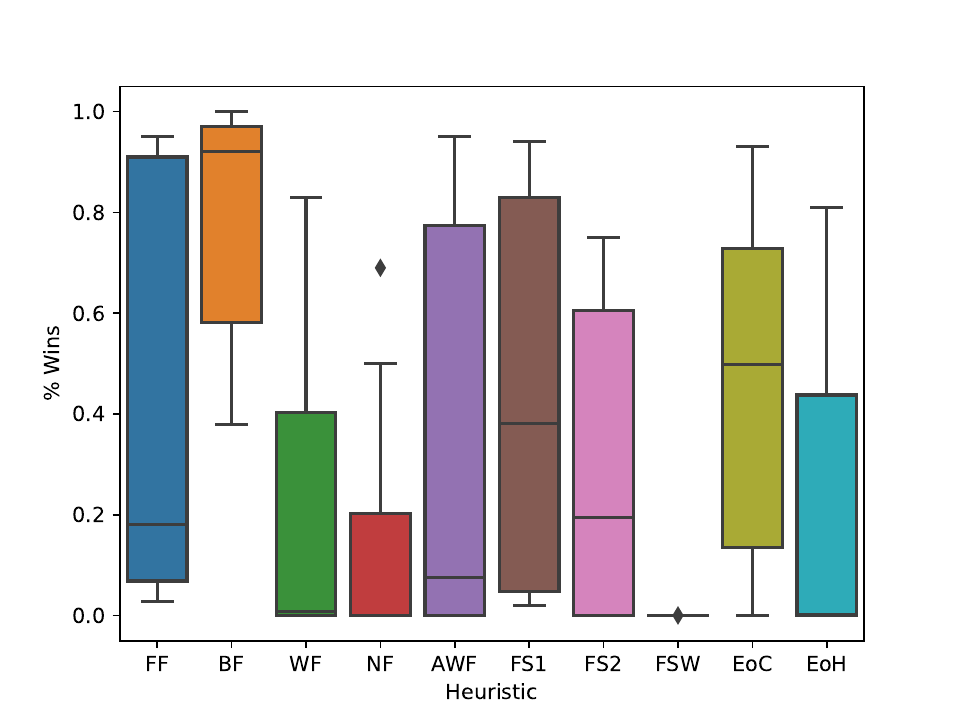}
        \caption{Wins on bins}
        \label{fig:boxplots_wins}
    \end{subfigure}
    
    \caption{\label{fig:independentBoxplots} Boxplots showing aggregated heuristic performance over all dataset for 2 metrics: average excess bins;  percentage of instances won.}
\end{figure}

Figure~\ref{fig:independentBoxplots} depicts the performance of each heuristic across the entire benchmark suite. It is clear  from Figure~\ref{fig:boxplots_bins} that $FSW$ has a very large interquartile range with respect to the AEB metric, as well as the poorest median. $FS2$ and $EoH$ have larger interquartile ranges and worse median values than their closest counterparts, $FS1$ and $EoC$ respectively. 
The larger interquartile ranges for AEB indicate that a heuristic does not generalise well, i.e. is a \textit{specialist} that performs well on a particular type of  dataset but delivers  poor performance on others. On the other hand, small ranges indicate heuristics are good generalists, i.e. work well across many datasets. Figure~\ref{fig:boxplots_wins} shows that the percentage  of instances in each dataset won by each heuristic varies widely. It is clear that $BF$ is an excellent generalist heuristic (median fraction of instances won per dataset $0.92$), outperforming all of the LLM-generated heuristics in this respect.  $EoC$ has the next best median value ($0.50$), winning more instances than the generalist hand-designed heuristic $FF$ (median 0.18). This is somewhat surprising given that the heuristic is evolved using Weibull datasets, and there is only one example of this type of dataset in the benchmark.

\section{Tuning LLM-Generated Heuristics}
As apparent in Figure~\ref{fig:allheuristics}, the  LLM-generated heuristics include several numeric constants, for example, used in condition statements or to scale a penalty score for a bin. To gain further insight into how robust the chosen parameters are, we attempt to tune each of these parameters for the five evolved heuristics. 

With the exception of $EoC$, tuning is performed using \textit{irace}~\cite{irace}\footnote{version 3.5} in its default setting using $5{,}000$ evaluations. For $EoC$, the entire search space is enumerated as there is only two parameters with small ranges of possible values.
Tuning is conducted on datasets containing $5$ instances per heuristic, using instances generated at random, drawn from the same distribution as the one on which the heuristics were trained. The performance metric used in tuning is the average excess of bins. Powers are integers tuned in the range $1-10$ to the exception of $FSW$ with range $1-8$ (causing errors otherwise). The addition constants in $EoH$ are real numbers between $0-10$. The initial penalty in $FS2$ is an integer between $50-10{,}000$ and the conditions in the IF clauses in $FS1$ are set to be between the previous condition and $100$.

Results are shown in Tables~\ref{tab:tuningSummary} and~\ref{tab:tuning}. The tuning process did not find any parameters for $EoC$ that improved on the default settings and hence this heuristic is not included in these tables. Table~\ref{tab:tuningSummary} shows the percentage improvement in the average excess bins  metric over the entire benchmark suite when  using the tuned heuristic compared to the untuned one. Negative values indicate a reduction in performance and vice versa. As a general remark, the evolved heuristics are robust: aggregating over the whole benchmark suite, tuning delivers small percentage improvements for $FS2$ and $FSW$, and small losses for $FS1$ and $EoH$.  Recall from Section~\ref{sec:BenchmarkResults} that both $FS2$ and $EoC$  obtain the best AEB on one dataset each, $FS1$ ties for 1st place with three hand-designed heuristics on one dataset, and  $EoH$ loses on every dataset. We suggest that in general these heuristics are so specialised to the specific datasets they were evolved on that no amount of tuning can address this.

\begin{table}
    \centering
        \caption{\textit{The first columns shows the percentage gain/loss in AEB averaged over the entire benchmark suite. The second column gives the percentage of the $12$ datasets for which an improvement was found. }}
    \label{tab:tuningSummary}
    \begin{tabular}{|c|c|c|}
    \hline
        Heuristic & Percentage Gain/Loss AEB & Percentage of datasets improved\\
        \hline
       FS1  & $-2.50$ & $57$ \\
       FS2 & $+0.63$ & $57$ \\
       FSW &  $+0.88$ & $67$\\
       EoH &$ -1.01$ & $0$ \\
       \hline
    \end{tabular}
\end{table}
\vspace{-1cm}
\begin{table}
\centering
\caption{\label{tab:tuning} Comparison of tuned and untuned evolved heuristics on datasets with the same distributions used in evolution.}
\subfloat[OR datasets]{
    \begin{tabular}{|c|c|c|c|c|}
    \hline
         &OR1& OR2 &OR3 & OR4 \\
         \hline
      FF &$6.42$ &  $6.45$ & $5.74$  & $5.23$ \\ 
      BF  & $5.81$ & $6.06$ & $5.37$ &  $4.94$ \\
      FS1 & $5.3 $& $\mathbf{4.19}$ & $\mathbf{3.11}$ & $\mathbf{2.47}$ \\
     \textit{ FS1 (T}) & $5.71$ & $6.06$ & $5.54$ & $5.16$ \\
      FS2 &  $\mathbf{5.2}$ & $5.12$ & $4.22$ & $3.84$   \\
      \textit{FS2 (T)}  & $5.4$ & $5.17$ & $4.37$ & $3.97$  \\
      \hline
    \end{tabular}
}
\quad
\subfloat[Weibull datasets]{\begin{tabular}{|c|c|c|c|}
\hline
        &  1K  &  5K &  10K\\
         \hline
      FF & $5.14$  & $4.46$ & $4.31$\\ 
      BF  & $4.63$ & $4.14 $& $4.00$\\
      FSW & $3.66 $& $\mathbf{0.69}$ & $\mathbf{0.37}$\\
      \textit{ FSW (T)}  & $3.52$ & 	$0.71$ &$ 0.38$ \\\
       EoH  & $2.18$ & $0.75$ & $0.56$ \\ 
       \textit{EoH (T)} &  $\mathbf{2.17} $& $0.73$ &\ $0.56$ \\ 
      \hline
\end{tabular}
}
\end{table}

\section{Evolving New Exemplar Instances}
In order to shed light on  which  heuristics work well on which type of instances, we generate new datasets, each of  which contains instances that are won by a single heuristic. 
That is, given a portfolio $\mathcal{P}$ of heuristics of size $n$, for each heuristic $h_i$ we evolve $100$ new instances in which $h_i$ outperforms the remaining $(n-1)$ heuristics in $\mathcal{P}$. 

Instances are evolved using a modified version of the EA described by Alissa {\em et. al.}~\cite{alissa2019algorithm}. In brief, given a  target heuristic $h_i$, a population of candidate instances is initialised, where the length of each instance is $n$ and item weights are randomly drawn from a uniform distribution between $(20,100)$. The EA evolves an ordering of items such that the performance of $h_i$ exceeds the performance of all other algorithms $\in \mathcal{P}$. A heuristic is considered to `win' an instance if the number of bins needed to pack all items is fewer than that resulting from applying any other heuristic in the portfolio. The EA in~\cite{alissa2019algorithm} uses the Falkanauer  metric~\cite{falkenauer1992genetic} as the objective function: we follow this approach to evolve instances as the metric has previously been shown to be useful in guiding the search towards `winning instances', however, we stop the search  as soon as an instance is found that uses fewer \textit{bins} than all the other heuristics in the portfolio. This contrasts to~\cite{alissa2019algorithm} where the goal was to maximise the difference in fitness between the target heuristic and the next best from $\mathcal{P}$.

The method was applied repeatedly to evolve $100$ instances won by each of the $10$ heuristics. No instances were found that were uniquely won by $NF$ therefore this heuristic does not appear in the following results.  To visualise the relationship between instances and winning heuristics, we use the Instance Space Analysis (ISA) methodology proposed in~\cite{smith2023instance}: this projects instances from a high-dimensional feature space into a 2-dimensional space that preserves linear trends in both feature and algorithm performance distributions, revealing 
insights into the strengths and weaknesses of algorithms across different regions of the instance space\footnote{The ISA method of~\cite{smith2023instance} obtains a 2d projection that encourages linear trends in performance \textit{and} feature distributions via a customised dimension reduction method using optimization to achieve the goal combined with a feature-selection method}.
In order to perform ISA, we consider online bin-packing instances as time-series and extract time-series features using the Python package \emph{tsfresh}\footnote{version 0.20.1}.
Given the large number of features extracted, $209$, we performed feature selection using recursive feature elimination from the scikit-learn package~\cite{scikit-learn}\footnote{version 1.2.2} to obtain $10$ features (all features and the selected ones can be found in supplementary materials).
As it is difficult to observe trends from $9$ heuristics (and $900$ instances) in the resulting visualisation, we provide multiple perspectives on the full instance space, each of which highlights subsets of heuristics.

\begin{figure}[h]
    \centering
    \begin{subfigure}{0.48\textwidth}
        \centering
        \includegraphics[width=\textwidth]{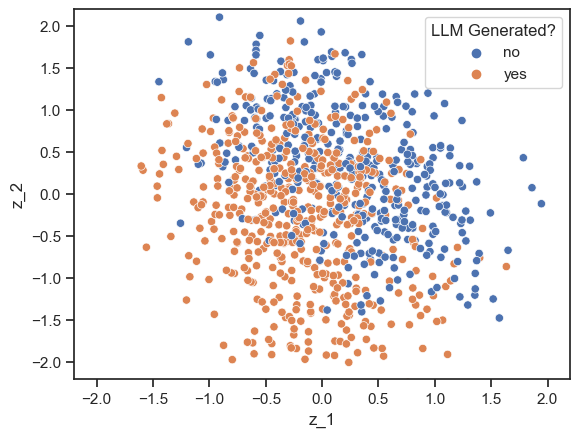}
        \caption{LLM vs hand-designed instances}
        \label{fig:ISA_llm}
    \end{subfigure}%
    \begin{subfigure}{0.48\textwidth}
        \centering
        \includegraphics[width=\textwidth]{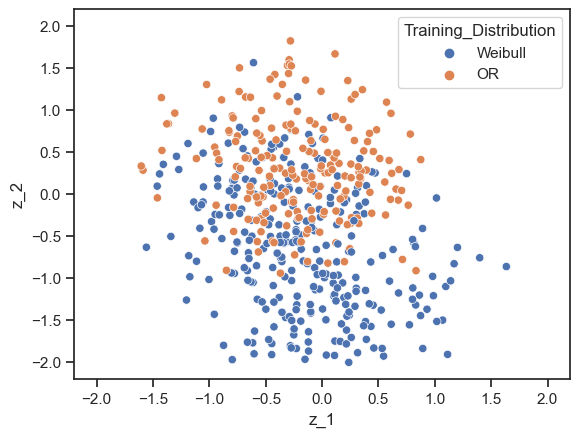}
        \caption{Weibull vs OR LLM heuristics}
        \label{fig:ISA_WOR}
    \end{subfigure}%
    \vskip\baselineskip
    \begin{subfigure}{0.48\textwidth}
        \centering
        \includegraphics[width=\textwidth]{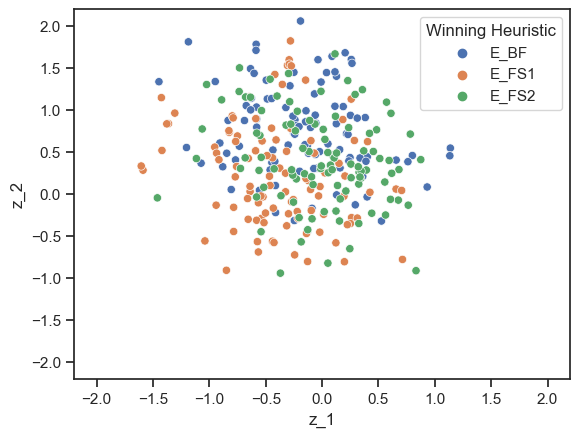}
        \caption{ORlib LLM-generated heuristics}
        \label{fig:ISA_OR}
    \end{subfigure}
     \begin{subfigure}{0.48\textwidth}
        \centering
        \includegraphics[width=\textwidth]{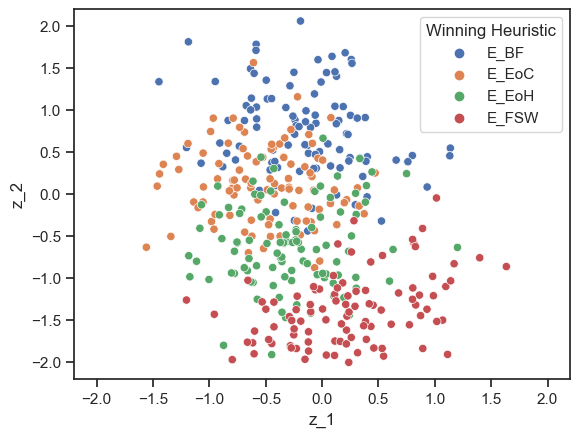}
        \caption{Weibull LLM-generated heuristics}
        \label{fig:ISA_W}
    \end{subfigure}
    \caption{\label{fig:ISA} Perspectives on the ISA projection of the instance space: (a) instances won by heuristics generated by an LLM vs hand-designed, (b) LLM heuristics trained on Weibull or OR distributions, (c) instances won by LLM heuristics originally evolved on datasets from ORLib with BF for comparison, and (d) instances won by LLM heuristics originally evolved on datasets with Weibull distributions, with BF for comparison.}
\end{figure}

From Figure~\ref{fig:ISA_llm}, we observe a relatively clear separation between areas of the space in which the LLM/Hand-designed heuristics are superior with some overlap in the middle.  Figure~\ref{fig:ISA_WOR} also shows that heuristics evolved using instances from Weibull distributions solve instances in a different part of the feature space to those evolved with OR instances,  highlighting the specialisation of heuristics to regions of the  instance space.  We previously observed that $FS1$ provides similar results to $BF$  (Section~\ref{sec:BenchmarkResults}); this is apparent  in Figure~\ref{fig:ISA_OR} where instances from all three heuristics occupy the same region of space. Finally, in  Figure~\ref{fig:ISA_W}, each of the heuristics evolved using Weibull distributions occupies  a distinct niche in the space; this niche is distinct from $BF$.

\begin{table}
\centering
\caption{\label{tab:gapToOpt} The fraction of evolved instances that were not won by each heuristic but are within ``\%  excess bins'' of the winning value. \textbf{1*}: the final row shows the result obtained on the independent datasets for comparison (for 1\% only).}
\begin{tabular}{|c|c|c|c|c|c|c|c|c|c|}
\hline
\% excess bins	& FF	& BF	& WF	& AWF & FS1	& FS2 & FSW	& EoC& EoH\\
\hline
10	&\textbf{1.00} &	\textbf{1.00} &	0.38 &	\textbf{1.00}	& \textbf{1.00} &	\textbf{1.00}	 &0.06	 &\textbf{1.00} &	0.83\\
5&	\textbf{1.00} &	\textbf{1.00}&	0.00 &	0.67&	\textbf{1.00}&	\textbf{1.00}&	0.00&	0.97	&0.24\\
2&	0.40&	0.53	&0.00	&0.09	&0.77&	\textbf{0.78}	&0.00	&0.45	&0.05\\
1&	0.09	&0.12&	0.00&	0.01&	0.20&	\textbf{0.28}	&0.00	&0.10	& 0.01\\
\hline
1* & \textit{0.20} &	\textit{0.40} & \textit{0.0007}&	\textit{0.004}	& \textit{0.30} &	\textit{0.27} &	\textit{0.0007}	&\textit{ 0.13}& \textit{0. 013} \\
\hline
\end{tabular}
\end{table}

To gain more insight into how a heuristic performs when evaluated on instances in regions of the instance space lying outside of the region that it is specialised to, we do the following: we apply each heuristic to the $800$ instances that it did not win, and record the fraction of these instances in which the heuristic produced a result that within $x\%$ of the winning number of bins. This provides an indication of how well the heuristic performs over the rest of the space. The result is summarised in Table~\ref{tab:gapToOpt}.  The LLM-generated heuristics $FS1$ and $FS2$ generalise very well, giving a result that is within $x=1\%$ of the best value for $20\%$ and $28\%$ of the instances respectively. With the exception of $FSW$ and $AWF$, all the heuristics generalise if $x=10\%$. $FSW$ has exceptionally poor performance on the 800 instances it was not evolved to win, showing it is very specialised to a single region of the space.  Note these results contrast to the heuristic ranking obtained using the independent benchmark datasets (Figure~\ref{fig:summary}): $FS2$ ranked 8th in this list, whereas on the evolved datasets it ranks 1st. The likely explanation lies in the distributions of item size per dataset: this varies widely in the independent datasets (Table~\ref{tab:datasets}) but in the evolved datasets is fixed to a single distribution.  These LLM heuristics therefore appear to generalise within this distribution but not across distributions.

\section{Conclusions}
Given the rapid rise in  production of new heuristics designed by frameworks that combine evolutionary algorithms and LLMs, a large-scale benchmarking exercise of these heuristics is opportune.

We conducted a rigorous study in the bin-packing domain on a large suite of benchmarks following typical practice in benchmarking in other combinatorial domains~\cite{mcmenemy2019rigorous,polyakovskiy2014comprehensive}, and used Instance Space Analysis~\cite{smith2023instance} to gain a more nuanced insight into the strengths and weaknesses of each of the heuristics evaluated.

Is the use of LLMs in generating new heuristics merely hype?
We find that the hand-designed heuristic $BF$ ranks as the best performing heuristic over the benchmark suite (average excess bins metric, AEB), also delivering performance within 1\% of the best heuristic in the portfolio on 40\% of instances it does not win. The LLM heuristic FS1 ranks second on the independent datasets. FS2 (LLM) generalises best on the \textit{evolved} datasets in delivering a result within 1\% of the best value on 28\% of instances it does not win, but recall that it was evolved using training data from the same distribution as used to generate  \textit{all} of the evolved instances. 
While the LLM heuristics clearly outperform hand-designed heuristics on the specific class of instance used to train them~\cite{romera2024mathematical,liu2024evolution}, most of them do not generalise to distributions that differ from their training examples (with the exception of $FS1$). Two LLM heuristics ($EoH$, $FSW$) appear very specialised, ranking 8th and 10th respectively on average AEB on the benchmark suite\footnote{The hand-designed heuristic $NF$ also does not generalise, ranking 9th}. In particular, $FSW$ has an AEB almost $20$ times that of the best-ranked heuristic $BF$ ($113.21$ vs $5.85$). $EoC$ is generally superior to $EoH$ in all our tests, while the opposite result is reported in \cite{liu2024evolution} on Weibull datasets.

There will of course always be a trade-off between specialist vs generalist heuristics. However, given  that the cost of discovering a new heuristic using an LLM is considerable (in terms of time and carbon emissions) it is important to quantify this trade-off. As the methods for designing new heuristics with LLMs improve, we hope that similar benchmarking exercises will enable the relative strengths and weaknesses of this exciting new research direction to  emerge.

\begin{credits}
\subsubsection{\ackname} Emma Hart and Quentin Renau are supported by funding from EPSRC award number: EP/V026534/1.

\clearpage
\subsubsection{\discintname}
The authors have no competing interests to declare that are
relevant to the content of this article. 
\end{credits}

\bibliographystyle{splncs04}
\bibliography{refs}

\begin{thebibliography}{10}
\providecommand{\url}[1]{\texttt{#1}}
\providecommand{\urlprefix}{URL }
\providecommand{\doi}[1]{https://doi.org/#1}

\bibitem{bpplib}
{BPPLIB} – {A} bin packing problem library. \url{https://site.unibo.it/operations-research/en/research/bpplib-a-bin-packing-problem-library}

\bibitem{alissa2019algorithm}
Alissa, M., Sim, K., Hart, E.: Algorithm selection using deep learning without feature extraction. In: Proceedings of the Genetic and Evolutionary Computation Conference. pp. 198--206 (2019)

\bibitem{angelopoulos2023online}
Angelopoulos, S., Kamali, S., Shadkami, K.: Online bin packing with predictions. Journal of Artificial Intelligence Research  \textbf{78},  1111--1141 (2023)

\bibitem{angelopGithub}
Angelopoulos, S., Kamali, S., Shadkami, K.: Binpackingpredictions. \url{https://github.com/shahink84/BinPackingPredictions/tree/main/Data/Benchmarks} (2024)

\bibitem{dataLLMBP}
Anonymous: {Beyond the Hype: Benchmarking LLM-Evolved Heuristics for Bin Packing - Data} (Nov 2024). \doi{10.5281/zenodo.14162744}, \url{https://doi.org/10.5281/zenodo.14162744}

\bibitem{beasley1990or}
Beasley, J.E.: Or-library: distributing test problems by electronic mail. Journal of the operational research society  \textbf{41}(11),  1069--1072 (1990)

\bibitem{bengio2021machine}
Bengio, Y., Lodi, A., Prouvost, A.: Machine learning for combinatorial optimization: a methodological tour d’horizon. European Journal of Operational Research  \textbf{290}(2),  405--421 (2021)

\bibitem{burke2013hyper}
Burke, E.K., Gendreau, M., Hyde, M., Kendall, G., Ochoa, G., {\"O}zcan, E., Qu, R.: Hyper-heuristics: A survey of the state of the art. Journal of the Operational Research Society  \textbf{64}(12),  1695--1724 (2013)

\bibitem{burke2012automating}
Burke, E.K., Hyde, M.R., Kendall, G., Woodward, J.: Automating the packing heuristic design process with genetic programming. Evolutionary computation  \textbf{20}(1),  63--89 (2012)

\bibitem{castineiras2012weibull}
Casti{\~n}eiras, I., De~Cauwer, M., O’Sullivan, B.: Weibull-based benchmarks for bin packing. In: International Conference on Principles and Practice of Constraint Programming. pp. 207--222. Springer (2012)

\bibitem{delorme2016bin}
Delorme, M., Iori, M., Martello, S.: Bin packing and cutting stock problems: Mathematical models and exact algorithms. European Journal of Operational Research  \textbf{255}(1),  1--20 (2016)

\bibitem{falkenauer1992genetic}
Falkenauer, E., Delchambre, A., et~al.: A genetic algorithm for bin packing and line balancing. In: ICRA. pp. 1186--1192. Citeseer (1992)

\bibitem{gschwind2016dual}
Gschwind, T., Irnich, S.: Dual inequalities for stabilized column generation revisited. INFORMS Journal on Computing  \textbf{28}(1),  175--194 (2016)

\bibitem{cocoJournal}
Hansen, N., Auger, A., Ros, R., Mersmann, O., Tusar, T., Brockhoff, D.: {COCO:} a platform for comparing continuous optimizers in a black-box setting. Optimization Methods and Software  \textbf{36}(1),  114--144 (2021). \doi{10.1080/10556788.2020.1808977}

\bibitem{HansenO01}
Hansen, N., Ostermeier, A.: Completely derandomized self-adaptation in evolution strategies. Evolutionary Computation  \textbf{9}(2),  159--195 (2001). \doi{10.1162/106365601750190398}

\bibitem{johnson1973near}
Johnson, D.S.: Near-optimal bin packing algorithms. Ph.D. thesis, Massachusetts Institute of Technology (1973)

\bibitem{liu2024evolution}
Liu, F., Xialiang, T., Yuan, M., Lin, X., Luo, F., Wang, Z., Lu, Z., Zhang, Q.: Evolution of heuristics: Towards efficient automatic algorithm design using large language model. In: Forty-first International Conference on Machine Learning (2024)

\bibitem{irace}
L{\'{o}}pez{-}Ib{\'{a}}{\~{n}}ez, M., Dubois-Lacoste, J., {Pérez C\'aceres}, L., Birattari, M., St{\"{u}}tzl, T.: The irace package: Iterated racing for automatic algorithm configuration. Operations Research Perspectives  \textbf{3},  43 -- 58 (2016)

\bibitem{martin2024automatic}
Mart{\'\i}n-Santamar{\'\i}a, R., L{\'o}pez-Ib{\'a}{\~n}ez, M., St{\"u}tzle, T., Colmenar, J.M.: On the automatic generation of metaheuristic algorithms for combinatorial optimization problems. European Journal of Operational Research  (2024)

\bibitem{mcmenemy2019rigorous}
McMenemy, P., Veerapen, N., Adair, J., Ochoa, G.: Rigorous performance analysis of state-of-the-art tsp heuristic solvers. In: European Conference on Evolutionary Computation in Combinatorial Optimization (Part of EvoStar). pp. 99--114. Springer (2019)

\bibitem{scikit-learn}
Pedregosa, F., Varoquaux, G., Gramfort, A., Michel, V., Thirion, B., Grisel, O., Blondel, M., Prettenhofer, P., Weiss, R., Dubourg, V., Vanderplas, J., Passos, A., Cournapeau, D., Brucher, M., Perrot, M., Duchesnay, E.: Scikit-learn: Machine learning in {P}ython. Journal of Machine Learning Research  \textbf{12},  2825--2830 (2011)

\bibitem{pluhacek2024using}
Pluhacek, M., Kovac, J., Viktorin, A., Janku, P., Kadavy, T., Senkerik, R.: Using llm for automatic evolvement of metaheuristics from swarm algorithm soma. In: Proceedings of the Genetic and Evolutionary Computation Conference Companion. pp. 2018--2022 (2024)

\bibitem{polyakovskiy2014comprehensive}
Polyakovskiy, S., Bonyadi, M.R., Wagner, M., Michalewicz, Z., Neumann, F.: A comprehensive benchmark set and heuristics for the traveling thief problem. In: Proceedings of the 2014 annual conference on genetic and evolutionary computation. pp. 477--484 (2014)

\bibitem{rinne2008weibull}
Rinne, H.: The Weibull distribution: a handbook. Chapman and Hall/CRC (2008)

\bibitem{romera2024mathematical}
Romera-Paredes, B., Barekatain, M., Novikov, A., Balog, M., Kumar, M.P., Dupont, E., Ruiz, F.J., Ellenberg, J.S., Wang, P., Fawzi, O., et~al.: Mathematical discoveries from program search with large language models. Nature  \textbf{625}(7995),  468--475 (2024)

\bibitem{sanchez2020systematic}
S{\'a}nchez, M., Cruz-Duarte, J.M., carlos Ort{\'\i}z-Bayliss, J., Ceballos, H., Terashima-Marin, H., Amaya, I.: A systematic review of hyper-heuristics on combinatorial optimization problems. IEEE Access  \textbf{8},  128068--128095 (2020)

\bibitem{scholl1997bison}
Scholl, A., Klein, R., J{\"u}rgens, C.: Bison: A fast hybrid procedure for exactly solving the one-dimensional bin packing problem. Computers \& Operations Research  \textbf{24}(7),  627--645 (1997)

\bibitem{smith2023instance}
Smith-Miles, K., Mu{\~n}oz, M.A.: Instance space analysis for algorithm testing: Methodology and software tools. ACM Computing Surveys  \textbf{55}(12),  1--31 (2023)

\bibitem{van2024llamea}
van Stein, N., B{\"a}ck, T.: Llamea: A large language model evolutionary algorithm for automatically generating metaheuristics. arXiv preprint arXiv:2405.20132  (2024)

\bibitem{van2024loop}
van Stein, N., Vermetten, D., B{\"a}ck, T.: In-the-loop hyper-parameter optimization for llm-based automated design of heuristics. arXiv preprint arXiv:2410.16309  (2024)

\bibitem{StornP97}
Storn, R., Price, K.: Differential evolution - {A} simple and efficient heuristic for global optimization over continuous spaces. Journal of Global Optimization  \textbf{11}(4),  341--359 (1997). \doi{10.1023/A:1008202821328}

\bibitem{djurasevic2022automated}
{\DJ}urasevi{\'c}, M., {\DJ}umi{\'c}, M.: Automated design of heuristics for the container relocation problem using genetic programming. Applied Soft Computing  \textbf{130},  109696 (2022)

\bibitem{wascher1996heuristics}
W{\"a}scher, G., Gau, T.: Heuristics for the integer one-dimensional cutting stock problem: A computational study. Operations-Research-Spektrum  \textbf{18},  131--144 (1996)

\bibitem{ye2024reevo}
Ye, H., Wang, J., Cao, Z., Berto, F., Hua, C., Kim, H., Park, J., Song, G.: Reevo: Large language models as hyper-heuristics with reflective evolution. arXiv preprint arXiv:2402.01145  (2024)

\bibitem{zhang2024understanding}
Zhang, R., Liu, F., Lin, X., Wang, Z., Lu, Z., Zhang, Q.: Understanding the importance of evolutionary search in automated heuristic design with large language models. In: International Conference on Parallel Problem Solving from Nature. pp. 185--202. Springer (2024)

\end{thebibliography}

\end{document}